# Experimental validation of Light CAble-Driven ELbow assisting device L-CADEL design

**M. A. Laribi[1], M. Ceccarelli[2], J. Sandoval[1], M. Bottin[3], G. Rosati[3]**

1. *Department GMSC, Pprime Institute, CNRS-University of Poitiers-ENSMA, UPR 3346, Chasseneuil Cedex, France*
2. *Department of Industrial Engineering, University of Rome Tor Vergata, Rome, Italy*
3. *Department of Industrial Engineering, University of Padova, Padua, Italy*

**Abstract**

This paper presents a new design of CADEL, a Cable-Driven Elbow assisting device, with lightweighting and control improvements. The new device design is appropriate to be more portable and user-oriented solution, presenting additional facilities with respect to the original design. One of potential benefits of improved portability can be envisaged in the possibility of house and hospital usage keeping social distancing while allowing rehabilitation treatments even during a pandemic spread. Specific attention has been devoted to design main mechatronic components by developing specific kinematics models. The design process includes an implementation of specific control hardware and software. The kinematic model of the new design is formulated and features are evaluated through numerical simulations and experimental tests. An evaluation from original design highlights the proposed improvements mainly in terms of comfort, portability and user-oriented operation.

## 1 Introduction

Assistive devices are those whose primary purpose is to maintain or improve an individual's functioning and independence to facilitate participation and to enhance overall well-being. Examples of assistive devices and technologies include wheelchairs, prostheses, hearings aids, visual aids, and specialized computer software and hardware that increase mobility, hearing, vision, or communication capacities [1].

*Corresponding author:* Med amine Laribi

**E-mail:** med.amine.laribi@univ-poitiers.fr

Motion assisting devices can be useful in rehabilitation therapies and training exercises of elderly people training [2]. Stroke patients require appropriate and persistent rehabilitation to recover lost abilities and to get back to their normal daily lives. Successful rehabilitation depends mainly on the level of damage in the brain and timing of rehabilitation. In general, patients receive rehabilitation sessions with therapist assistance as pointed out for example in [3]. The number as well as frequency of session are limited due the availability of therapists. To cope with this problem, rehabilitation support system that allows patients to carry out rehabilitation exercises by themselves becomes essential as pointed out for example in [4]. This kind of rehabilitation therapy that can be performed by patients at home is called self-controlled rehabilitation, [5][6] and requires system which can find a large market.

Many systems for rehabilitation have been proposed in literature as reported for example in [7][8]. In particularly Kerbs et al. proposed a robotic system to assist, enhance, quantify, and document neurorehabilitation [9]. Reinkensmeyer et al. provide a telerehabilitation system for arm and hand therapy following a stroke [10]. While Masiero et al. proposed a cable-suspended robotic device for rehabilitation of the proximal upper limb in patients at bedside when in acute phase after stroke, [11]. Volosyak et al. present improved version of rehabilitation robot FRIEND-I with the use of smart devices, camera systems, a human-like robot arm with 7-joint kinematics [12]. Mahoney et al. present a robotic therapy device, called ARCMIME, with potential to improve rehabilitation outcomes significantly for individuals, who have upper limb impairments due to stroke and other brain injuries [13]. More recently, Pang et al. proposed a 6-degree-of-freedom upper limb rehabilitation robot via using ropes as the mainstay and driving by the "rope + toothed belt" [14]. All these systems have proven to be valuable tools in the recovery of arm function, but they seem not suitable for patient self-controlled rehabilitation therapy. Arm rehabilitation therapy with robot aids can be performed by patient in self-controlled rehabilitation scheme.

In addition, physiotherapists commonly help elderly individuals with problems such as reduced mobility, poor balance, muscle weakness, and decreased independence as indicated for example in [15][16]. Rehabilitation plays an important part also in treating memory disorders [17]. Good results are achieved when rehabilitation takes place in an environment that is familiar to a patient, such as home or a nursing house, and when it is started at the earliest stage possible. Elderly rehabilitation and exercises may include: exercising to increase muscle strength and balance, and to reduce risk of falling; stretching to prevent soft tissue shortening and to increase range of movements and activities for better fitness levels as pointed out in [18].

On other hand, existing robotic systems present various architectures and in the context of this paper cable driven parallel robots (CDPRs) are ones of more suitable architectures. CDPRs also known as wire-driven robots are parallel manipulators with cables instead of rigid links. This robot structure consists of a base frame, a moving-platform and a set of cables connecting in parallel the moving platform to the base frame [19]. CDPRs are well-known for their advantageous performance over classical parallel robots in terms of large translational workspace, reconfigurability, large payload capacity and high dynamic performance [20]. In literature several examples have been proposed. Krebs et al. present a poly-articulated system, called MIT-MANUS robot [21], that is dedicated to shoulder and elbow rehabilitation in stroke patients. Beer et al. developed a Multi-Axis Cartesian-based Arm Rehabilitation Machine (MACARM) that is based on CDPR architecture for upper limp rehabilitation [22]. The proposed solution with 6 active DOFs suffers from a coupling issue between the end-effector position and orientation. Rosati et al. presented the design of MariBot, a wire-based robot, starting from the NeReBot experience previously developed [23]. Thanks to mechanical design improvement, MariBot resulted much lighter and less cumbersome than NeReBot.

CADEL (Cable-Driven Elbow assisting device) design, whose original design is reported in [24][25], is presented in this paper as improved solution based on the cable-driven philosophy, which makes the robot intrinsically safe. This new device is adapted to patient-controller therapeutic exercises for self-controlled rehabilitation type. CADEL is dedicated to elbow motion assistance and presents a portability advantage which allows a patient to perform motion exercises at home.

The paper is organized as follows: Section 2 deals with a description of the biomechanics of elbow and the design requirements of the rehabilitation device. Section 3 deals with a discussion of existing solutions and open issues in arm rehabilitation and particularly in motion assistance of the elbow articulation. Section 4 presents a description, a kinematic model as well as a performance characterization of CADEL, (the CAble-Driven for ELbow assisting device), versions with the obtained results, their analysis and improvements. In section 5, conclusions summarize the paper content with future work considerations.

## 2 Upper limb joints and motion characterization

Motion of the human body is a complex nervous and musculoskeletal systems performing a variety of interaction with the environment. Dysfunction of one part of this complex system may yields in abnormal motion that may trouble an individual's ability during daily living tasks. Understanding the models of muscle activation and joint motion is essential for simulating an healthy human operation.

Hand function is not limited to grasp and sensation but also the capability of the upper limb to move freely in 3D space. An individual's capacity to perform essential activities of daily living is subject to hand function, thus any disfunction of upper limb mobility may be leads to a loss of independence.

The loss of functional upper limb range of motion can be treated through rehabilitation exercises which allow to increase and recover joint range. The rehabilitation can be a solution to provide alternate motion strategies, or to provide replacement or assistive devices that compensate the lost range and/or function as indicated for example in [26].

From an anatomical point of view as shown on Figure 1, the upper limb consists of the arm (the upper arm), the forearm (the lower arm), and the hand. The arm consists of a single bone, the humerus. The forearm consists of two bones, the ulna and radius. The hand consists of 27 bones, which are grouped into the phalanges, metacarpals, and carpals [27].

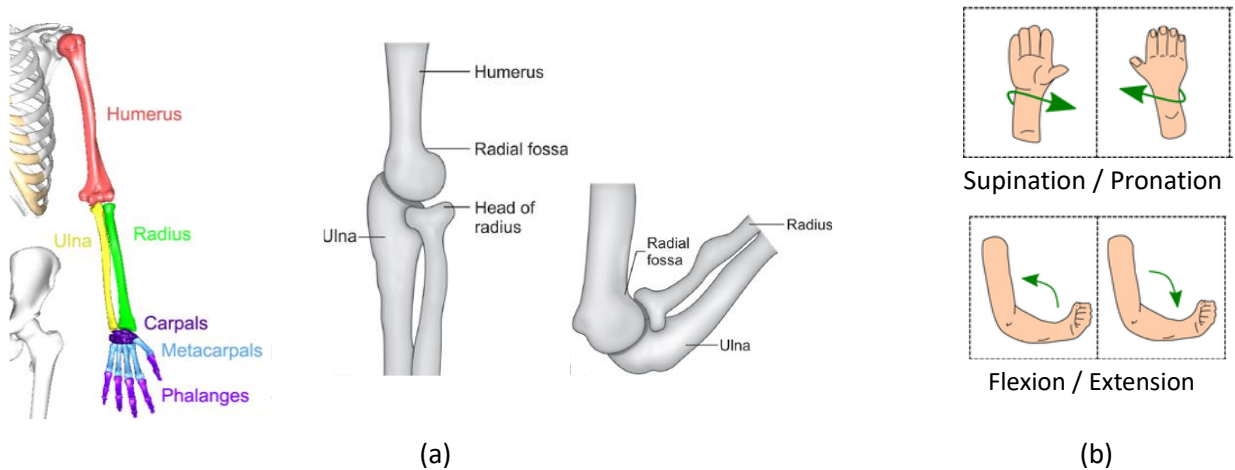

(a)            (b)

**Figure 1. Arm characteristics: a) bone structure; b) motion variety**

The human upper limb has three complex articulations; the shoulder, the elbow and the wrist. The forearm joint can be described as a complex joint due to the possible internal movements. The radius and ulna that interact together distally with the hand [1]. In Biomechanics, the forearm is commonly considered as a single segment. In general, the elbow and the wrist are modelled as universal joints where the pronosupination is often reported at the elbow, [28][29]. In this work, being dedicated to the elbow motion assistance, covers one degrees of freedom (DoF) (flexion / extension) and one DoF for the forearm joint. Accordingly, the kinematic model given on the Figure 2 is considered for the elbow and the forearm [30]. This model is adopted in the development of the assisting device. The main purpose of an assisting device is not only to provide efficient motion assistance to the human limbs, but also to guarantee the safety and the comfort of a user. That is why matching the human upper limp anatomy is one of the most important criteria for a device design.

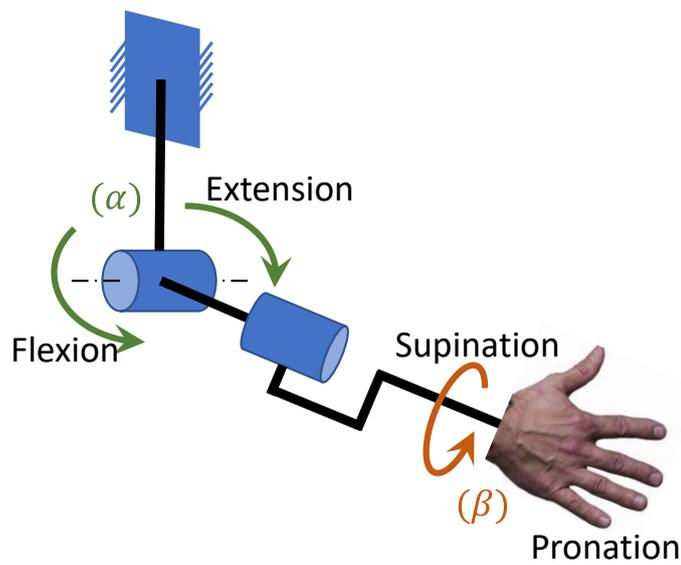

Figure 2. A kinematic model for elbow and forearm.

An estimation of range of motion (RoM) for each joint of the upper limb can be obtained using a suitable setup of motion capture system. The forearm flexion/extension motion as well as the forearm pronation/supination motion can be described respectively by the angle $\alpha$ and the angle $\beta$, as shown on Figure 2. Table 1 gives the average range of the elbow as well as forearm joints of the human arm.

**Table 1. Human arm characteristics, [30]**

|  | Elbow<br>Flexion/ Extension | Forearm<br>Supination / Pronation |
|---|---|---|
| Range of motion | $\pm 60°$ | $\pm 50°$ |

## 3  Existing rehabilitation solution

Robots have been steadily gaining importance especially in industrial applications, but also entering other fields such as medicine, entertainment, construction, space or personal use, like the example in [9][10]. Those robots performing useful tasks for humans or equipment excluding industrial applications fall into a different group, the so-called service robots [11]. The role of robots in movement disorders and post-stroke patients' rehabilitation has been investigated intensively as pointed out for example in [12]. In fact, studies have highlighted a need to provide rehabilitation procedures with some key features for the effectiveness of repetitive performance and a rehabilitation strategy. It is also important that a patient can perform an assigned task continuously and without dependence on a physical presence of an assistant. The use of robotic devices for rehabilitation is of particular importance as pointed out in [7][13]. Robotic rehabilitation is an effective strategy for at least partial recovery of the lost functions and robot devices are good solutions for home exercises of elderly people. Robot medical device allows precise and repeatable movements, and evaluate recovery and improvement in running the motor task and activation task muscle during the motion exercises as pointed out in [31]. The repetition of a same gesture leads to improvement in function recovery provided that the use of robotic technique is frequent and prolonged. Robotic rehabilitation features many benefits, such as repeatability, autonomy during task execution, chance of studying and validating new protocols based on new motion laws, introduction of different strategies to grant faster and better recovery as pointed out in [13].

Table 1 presents a list of the main existing rehabilitation devices with robot structures with cables. For each device the number of DoFs (degree of freedoms), joints, actuation type as well as power transmission are given as an indication of main characteristics.

**Table 2. Existing upper-limb exoskeleton robots for motion assistance.**

| Design | Joint | DoF | Actuators | Power Transmission |
|---|---|---|---|---|
| ARMin [14] | shoulder, elbow, wrist | 7 | DC Brushed Motors, Maxon series RE | cable, gear |
| CAREX [15] | shoulder, elbow | 4 | Brushless Motors, Maxon EC 45 | cable |
| EXO-7 [16] | shoulder, elbow, wrist | 7 | DC Brushed Motors, Maxon | cable |
| L-EXOS [17] | shoulder, elbow | 4 | AC Motors, VERNITRON 3730V-115 | cable, gear |
| IntelliArm [18] | shoulder, elbow, wrist, hand | 8 + (2) | Electric motors | cable, gear |
| MEDARM [19] | shoulder, elbow | 6 | Electric motors | cable, gear |
| MGA [21] | shoulder, elbow | 5 | DC Brushless Motors | gear |
| SUEFUL-7 [22] | shoulder, elbow, wrist | 7 + (1) | Harmonic Drive series RH/RHS | cable, gear |
| MariBot [23] | shoulder, elbow | 3 + (2) | Direct drive pulley-motor system | cable |

Open issues related to motion assistance of the elbow joint can be recognized as a specific application in arm rehabilitation. With reference to Figure 3, aspects relating to interaction with patients and interaction with medical operators can be recognized as fundamental. These aspects may have an important impact not only in a concrete application of motion assistance but even in the design of the structure and in the regulation of its functionality as well as in the choice of materials and production processes that can lead to suitable devices for both motion assistance needs and for satisfaction of users. In particular, the following aspects are emphasized in Figure 3, as differentiated but shared from the point of view of patients and medical operators.

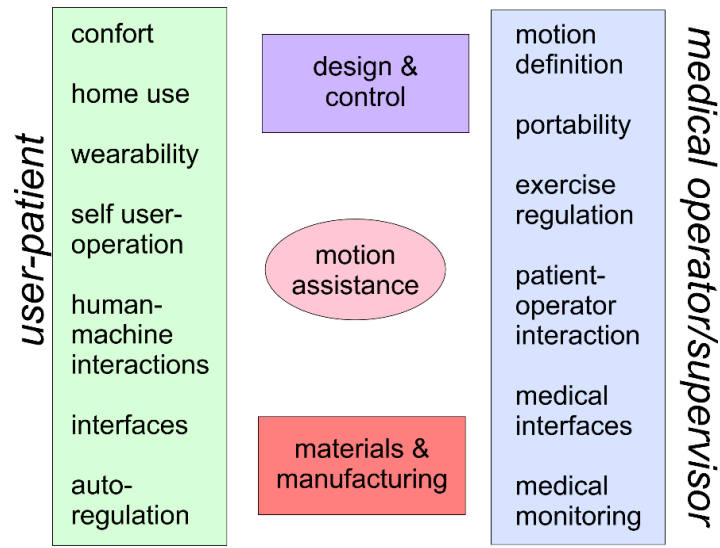

**Figure 3. Open issues on motion assistance of elbow articulation.**

As per the patient side, reference can be made mainly to:

**Comfort**: User comfort can be a requirement that gives constraints and requirements that are often not considered for the functional medical purposes of a motion assistance device. The comfort in wearing and functionality of the device require considerations not only in design and functional problems but also in man-machine interactions and psychological effects on user.

**Home use**: The solution of a portable device that can be used at home is a problem both for the design and for the operation of a motion assisting device that must be managed directly by a user in an environment such as the home environment that is not specific for applications of medical devices and therefore it may require particular attention both in terms of sizing and functionality as well as interactivity with a user, bearing in mind requirements on medical and environmental health and hygiene.

**Wearability**: Wearability is linked both to comfort and to the possibility of home use and it is also an important aspect for patient user acceptance, requiring not only comfort and transportability but also aspects of interaction with patient's arm and the ease of application and use of the device. These problems certainly influence the design and operation of the device, also requesting specific sensors to define an optimal solution.

**Self-user operation**: Autonomy in managing the device for motion assistance is a fundamental aspect for determining a device that can be managed directly by a patient-user considering both learning and handling the functionality of the system as well as its regulation and monitoring during motion exercise. This aspect includes also device quality in terms of training and ease of device management.

**Human-machine interactions**: The interactions between the limb and the device determine problems and characteristics of the solution for motion assistance that can be understood as requirements and constraints but also as indications for the design and functionality of the device at adequate levels both in terms of movements and actions between the device and the user's limb.

**Interfaces**: The interfaces that can be required for system operation are not only at the level of movement adjustment for the exercise by the user but there may be also interfaces for both motion and clinical monitoring. Desirable are interfaces that can facilitate the understand of the device by a user and its management, such as the supervision of the system via smartphone.

**Auto-regulation**: the regulation and therefore the control of the functionality of the device in the exercise of motion assistance require supervision algorithms that can be independent of the user's reaction and indeed it is required that the device can have self-regulation to ensure levels of operation as also per the user's personal safety.

**Safety**: User safety is an essential aspect in these motion assisting devices where a machine interacts with human body and the needs to ensure high reliability levels impose design choices and controlled functional procedures in every aspect of risk which, however, must also be shared with aware by users and medical operators.

As per the medical operator side, reference can be made mainly to:

**Motion definition**: The definition of the characteristics of motion in the exercise of motor assistance requires medical clinical knowledge of a patient by a medical operator and the possibility of applying this exercise adequately with the device available. The definition of motion exercise requires synergy and integration of the device's motion skills with the physiotherapy requests that are identified by a medical operator specifically for a user-patient.

**Portability**: Portability is a specific requirement of these of motion assisting devices that makes them very suitable to the needs of patients and to the possibility for physiotherapists to adapt motion assistance with specific patient requirements. This portability feature, however, requires considerable attention in terms of lightness in the device as well as its easy functionality, easy adaptability to patient anatomy and operation management by a patient himself with few elements supporting the device.

**Exercise regulation**: Exercise regulation is a very important aspect in motion assistance that must be possible both by a user and by a medical operator supervising a medical therapy. Therefore, regulation and control of the motion exercise and of the sensorizations that are connected to it must be at the levels of the functionality that can be managed by a patient as well as useful for monitoring and remote monitoring by a medical supervisor therefore requiring both an autonomy of control and a control open to contextual and temporary needs of the patient and of the medical supervisor.

**Patient-operator interaction**: an interaction between a patient and a medical operator during the motion assisted exercise must be provided in order that they both can react to the responses to the exercise and make necessary adjustments as well as obtain data on the achieved results.

**Medical interfaces**: Medical interfaces are indispensable aids for a motion assistance with regard to clinical and medical parameters during and in response to motion exercise. Such medical interfaces in terms of both sensors and systems as well as a visualization interaction can be useful complements and they must be integrated into the structure and functionality of the device so that both the design and the operation of the device need to consider these interfaces as part of the system yet.

**Medical monitoring**: Medical monitoring can be considered indispensable not only for the supervision and evaluation of the motion exercise in assistance with regard to its effectiveness but also for the possibility of giving timely indications to a patient-user during and in anticipation of an exercise. Medical monitoring systems can be composed of the same sensors as the medical interfaces but also special devices that can allow remote monitoring with video data transmission solutions.

**Safety**: aspects of safety, including medical ones, are certainly to be considered as a priority in the applications of devices for motion assistance and from a medical point of view they also require non-technical considerations that can continuously identify new problems and new solutions in terms of both technical and medical devices.

The above-mentioned aspects can be considered also as an overview of the open issues and the related problems that should be addressed in the development of a technical-medical device for motion assistance in rehabilitation, mobility or training exercises. The aspects considered above as well as others that can be identified on the basis of specific applications, are the bases for both technical and medical considerations that must be taken into account separately and in integrated way in the design and functionality of a system for motion assistance that will be satisfactory for the three levels of use, namely patient, technical side, and medical application.

## 4 CADEL versions and improvements

A specific solution for the motion assistance of the elbow was conceived considering the considerations previously discussed and with reference to the conceptual solution summarized in Figure 4. In particular, Figure 4-(a) shows a scheme for a solution with the main characteristics for a device that is portable, comfortable and easy to use for a user responding to requirements of suitable mobility and support in force transmission while keeping in mind the expectations in biomechanical and medical terms. Figure 4 (b) represents the conceptual solution that was conceived with the use of a parallel architecture system with cables so that it is wearable and can replicate and help the functionality of the neuromuscular structure for the functionality of the elbow joint, [32]. The original conceptual idea is in fact conceived with two ring platforms that can be worn on the arm and the forearm with a relative movement that is operated by means of cables whose driving servo motors are on board the ring platform on the arm, Figures 4 and 5.

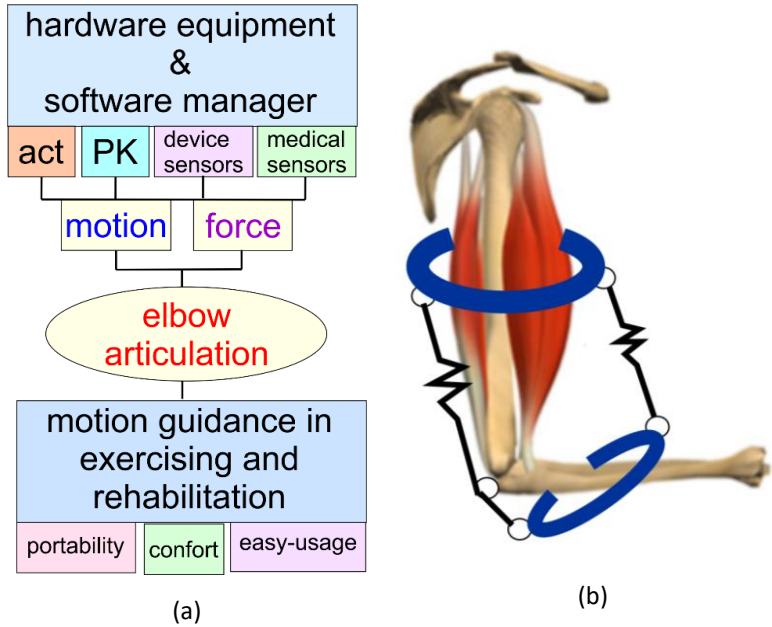

**Figure 4.** Conceptual design of CADEL: a) a block diagram; b) a functional scheme

The original CADEL solution as reported in the publications of 2017 [7][34] was developed according to the kinematic scheme of Figure 5 in which it can be noted that the four servomotors for actuating the cables are installed on the platform ring of the arm. Forearm ring platform is equipped with a shaft that is designed to accommodate the forearm and at the same time to facilitate the connection of the corresponding cable. The kinematic structure is characterized by the four cables that work independently to obtain the movement of flexion and extension and rotation and torsion that are represented respectively by the angle $\alpha$ and the angle $\beta$.

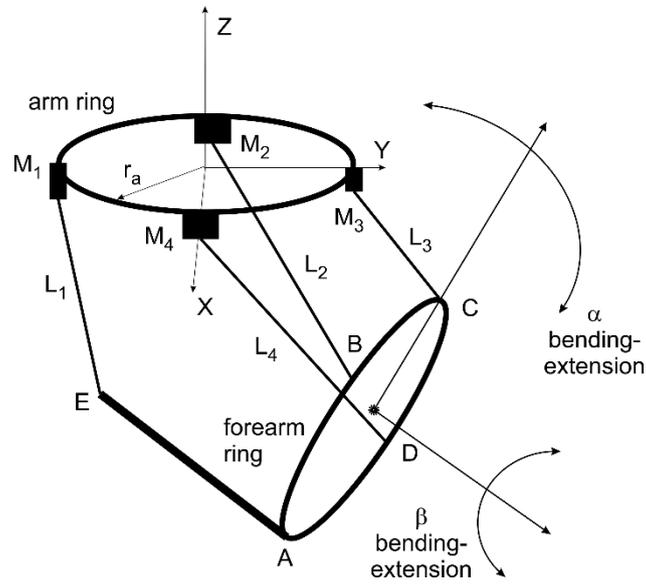

**Figure 5. A Scheme of kinematic design of CADEL with design parameters**

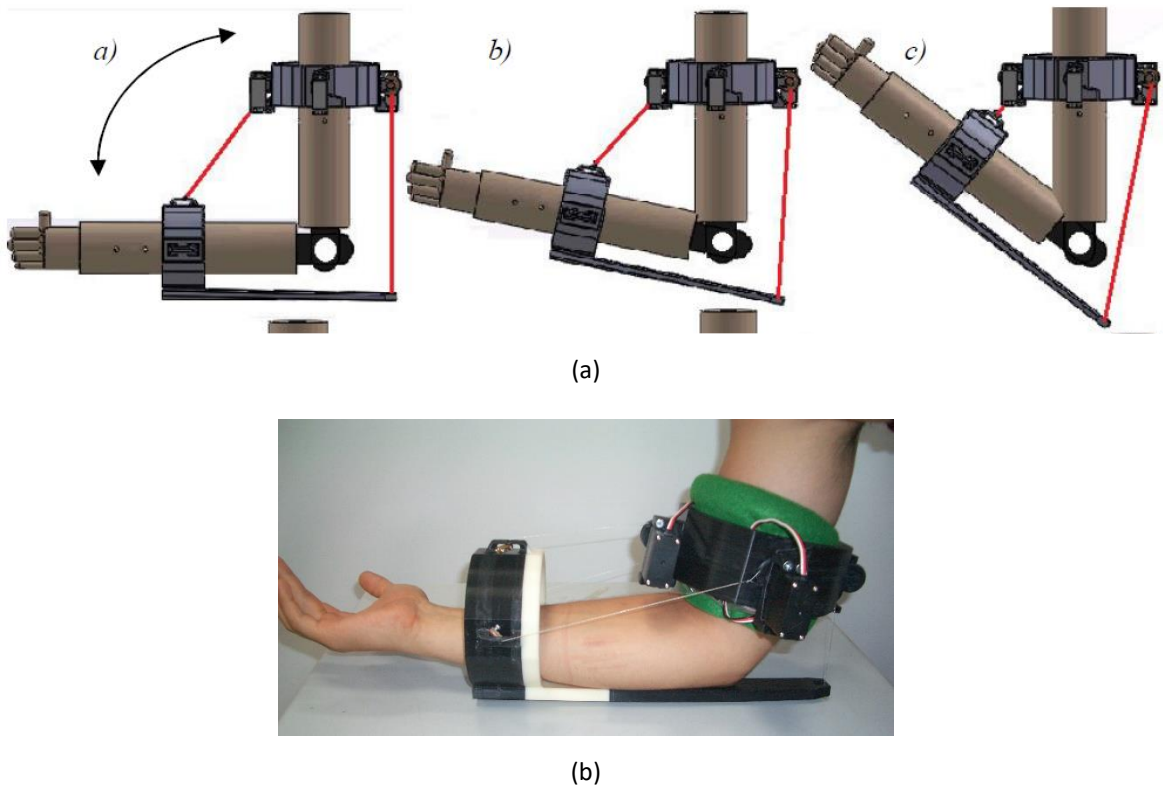

**Figure 6. Original CADEL design in 2017 [33]: a) a simulation for performance evaluation; b) the first prototype**

Figure 6 shows the design characteristics of the first solution that was developed following dynamic simulations as shown in Figure 6-(a) for definition and characterization of the functionality with an optimization of the main components and it was validated with the construction of a first prototype as in Figure 6-(b) built using 3D printer-produced platform rings and low-cost commercial components. The experimentation with the first prototype is summarized with the illustrations of Figures 7 and 8 with characteristic results in Figure 9 in tests with and without load on the arm that have highlighted, in addition to the validity of the system in terms of motion support to exercise and of the rehabilitation of the elbow, also the limits of practical utility both in quantitative and qualitative terms. The first prototype of the CADEL device showed deficiencies in terms of comfort for a user and limited easy wearing due essentially to the bulky mechanical structure of the two ring platforms and the forearm support rod. The numerical results in the examples of Figure 9 show the characteristics with numerical values suitable for adequate motion assistance in terms of support and smoothness of movement as well as a low operating power of the device which prove the feasibility of the device in its portability and user-oriented application.

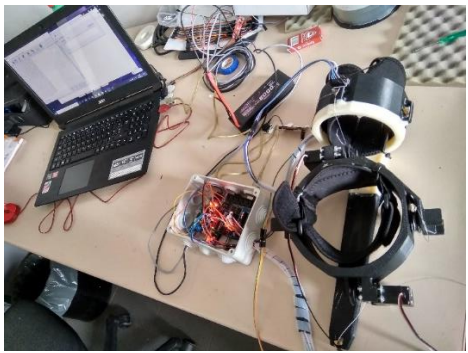 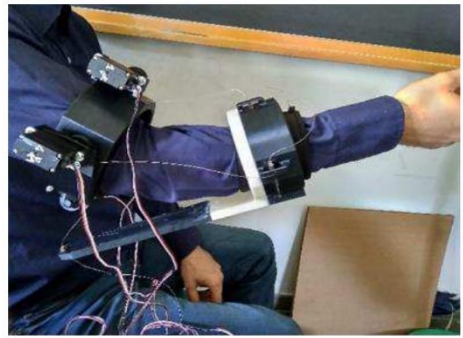

(a) (b)

**Figure 7. CADEL first prototype for testing in Rome [34]: a) lab setup; b) testing when worn on an arm**

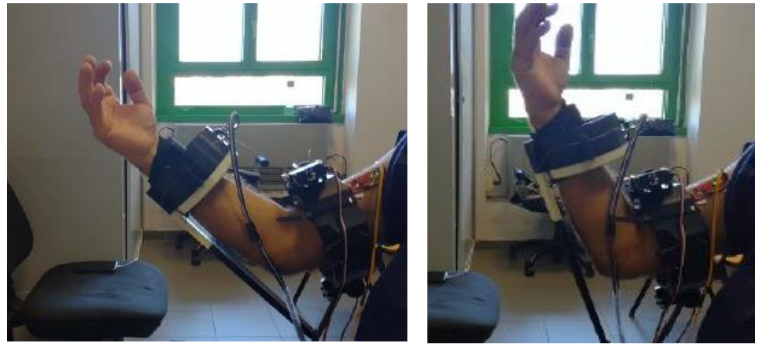

(a)

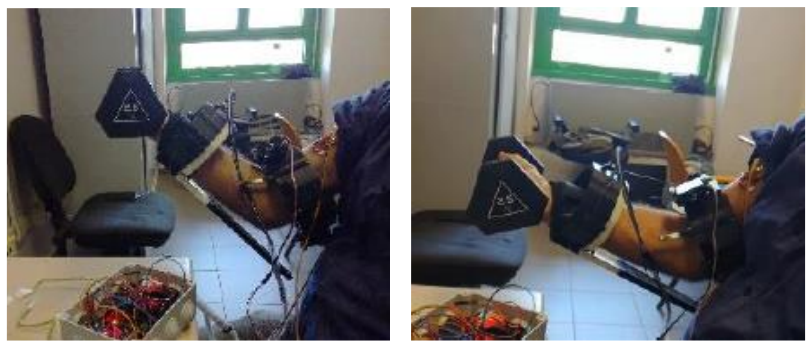

(b)

**Figure 8. A snapshot lab testing in Rome with prototype in Fig.7: a) worn without load; b) worn with load**

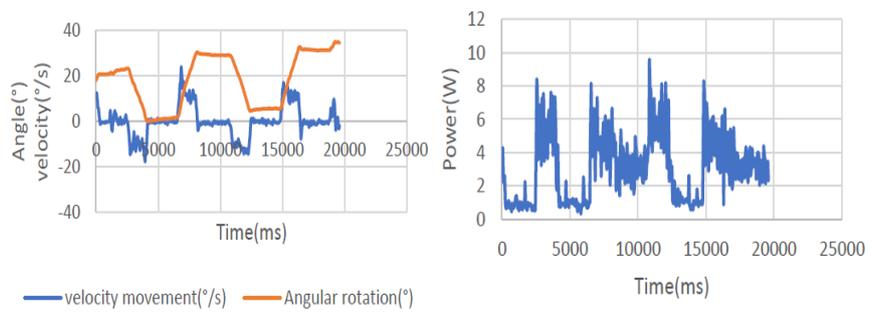

(a)

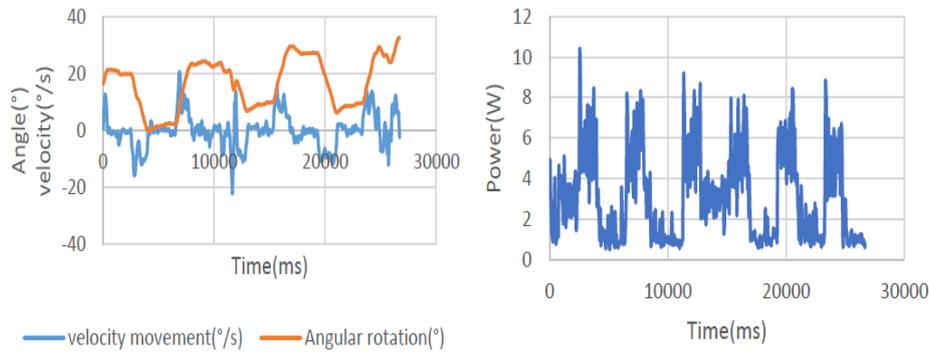

(b)

**Figure 9. Results of test like the one in Fig 8 in terms of arm rotation angle and power consumption: a) without load as in Fig 8 a): with load as in Fig. 8 b)**

The laboratory experiences made with the first prototype allowed to highlight problems that required improvements both in structural and functional terms according to two different research lines. The first one was focused on structural improvements as in the CADEL.3 version represented in Figure 10 and then on a validation of the expected functionalities and consequent improvements as reported in the experimental results briefly illustrated in Figure 11, [35]. In particular, the most significant structural aspects of the CADEL.3 prototype in Figure 10 can be summarized in the new design of the ring platforms with leaner and lighter solutions together with an inflatable interface for a simple and more efficient wearing together with avoiding the forearm accommodation rod by realizing a sort of jacket which at the elbow has a guide pulley for the cable corresponding to that actuation part of the device. The four-cable structure was maintained, however, leaving the lateral cables with the primary function of maintaining the assisted movement in the sagittal plane of the arm as managed by the other two cables in flexion and extension. The resizing of the rings has also made it possible to optimize the positioning of the servomotors for actuating the cables and to reduce their size according to the power necessary for the intended motion. The prototype created was tested objectively with a laboratory solution as in Figure 10-(a) and subsequently with tests on a human operator, Figure 10-(b), in order to verify its effective functionality with the expected improvements in comfort and efficiency of use. These characteristics have been numerically evaluated during experimental tests with results reported in [35][36] and summarized by the diagrams in Figure 11 to indicate both the efficiency of actuation for different load levels on the arm, Figure 11-(a), and the limited value of energy required for its operation, Figure 11-(b).

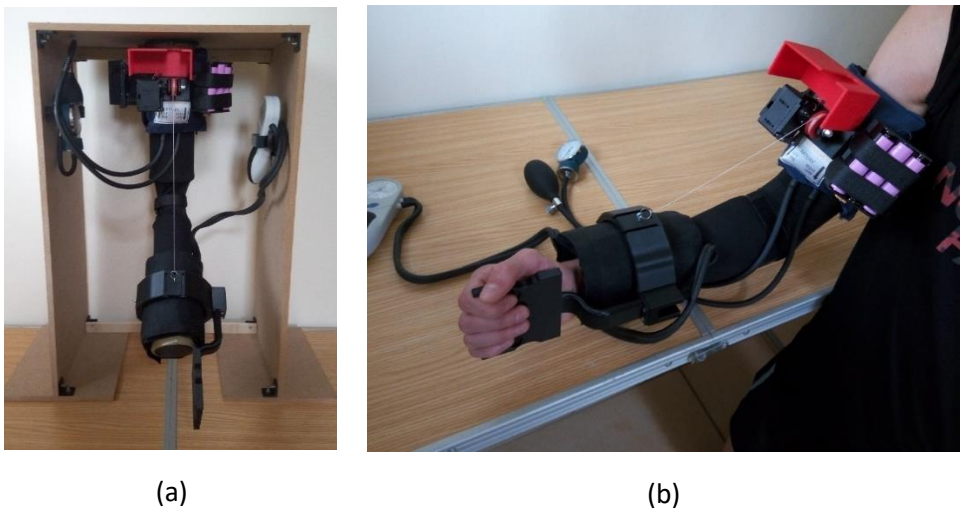

(a) (b)

**Figure 10. CADEL.3 prototype at Lab of Robotics in Padova in 2020: a) the lab setup; b) when worn for testing.**

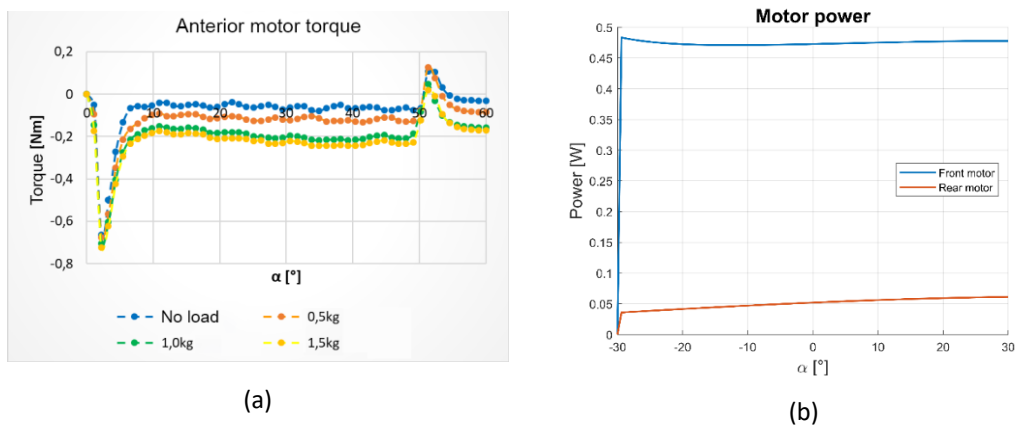

(a) (b)

**Figure 11. Test results with Padova prototype in Fig 10 for a flexion/bending exercise: a) motor torque on anterior motor b) power consumption.**

The second approach was mainly focused on improvements related to the efficiency and functionality of the device also considering what was developed in the CADEL.3 version. This attention led to the solution represented in Figure 12-(a), with a CAD solution that has been simulated for a numerical characterization before making the prototype in Figure 12-(b) with the essential characteristics that can be summarized in the new two ring platforms equipped with an inflatable interface for comfort and stable wearing and in the realization of motion assistance in the sagittal plane for extension and flexion of the forearm with only two cables whose servomotors are movable to accompany the rotary motion imposed by the rotation of the forearm. This solution has made possible to further lighten the mechanical design of the device, to reduce the energy required for its operation, and to increase its comfort both during wearing and when operating in the motion assistance.

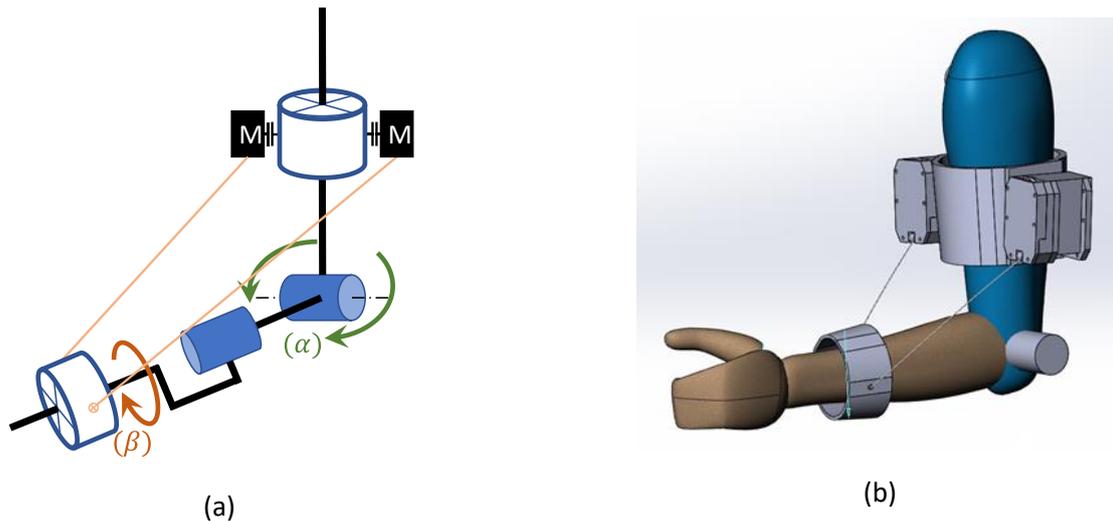

(a)  (b)

**Figure 12. A CAD model for a light CADEL design in Poitiers in 2020: a) conceptual design b) a model for performances simulation**

The prototype of light CADEL design, shown in Fig. 13, was tested objectively with a laboratory solution in order to verify its functionality with the expected improvements in easiness of use and efficiency. These characteristics have been numerically evaluated during experimental tests with results reported in Figures 15 and 16. The prototype has been fixed to an aluminum structure simulating a human arm, as depicted in Fig. 14.

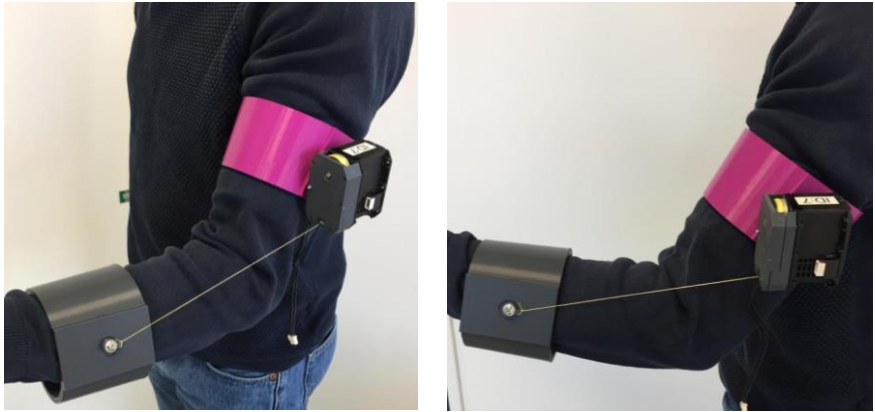

**Figure 13. A prototype of light CADEL design in Fig. 12**

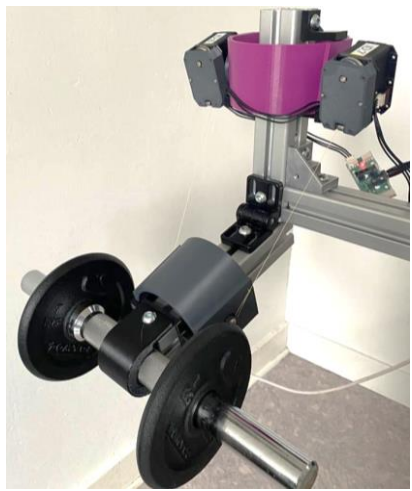

**Figure 14. Test bench for the light CADEL prototype**

A position control law has been used to execute a flexion/bending trajectory, with a sampling time of 6ms. In order to verify the performance of the prototype, a desired trajectory has been executed for different loads: 0,5 kg, 1,0 kg, 1,5 kg and 2,5 kg. Figure 15 shows the angular trajectory imposed by the light CADEL to the elbow of the test bench, having defined the zero position when the forearm is in the horizontal plane.

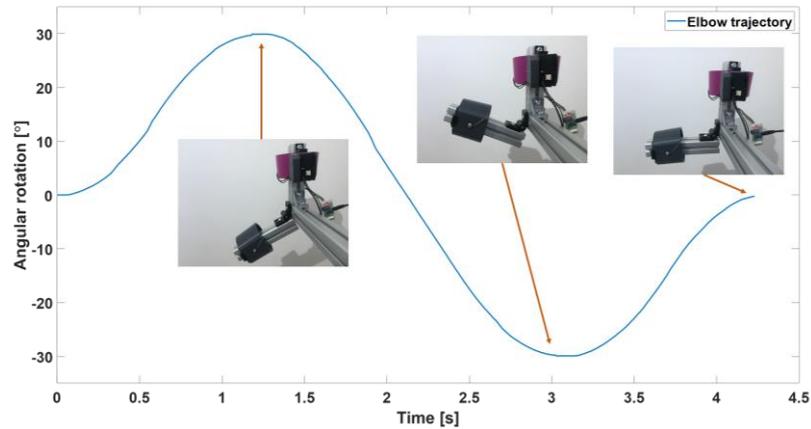

**Figure 15. Test results with the light CADEL prototype for a flexion/bending exercise: executed elbow trajectory**

Since the load of the fore arm is distributed equitably between the two cables, the produced torque of the two motors is similar. Therefore, we only present in this paper the response of the motor placed at the right side. Figure 16 shows the torque produced by the right-side motor for the different loads.

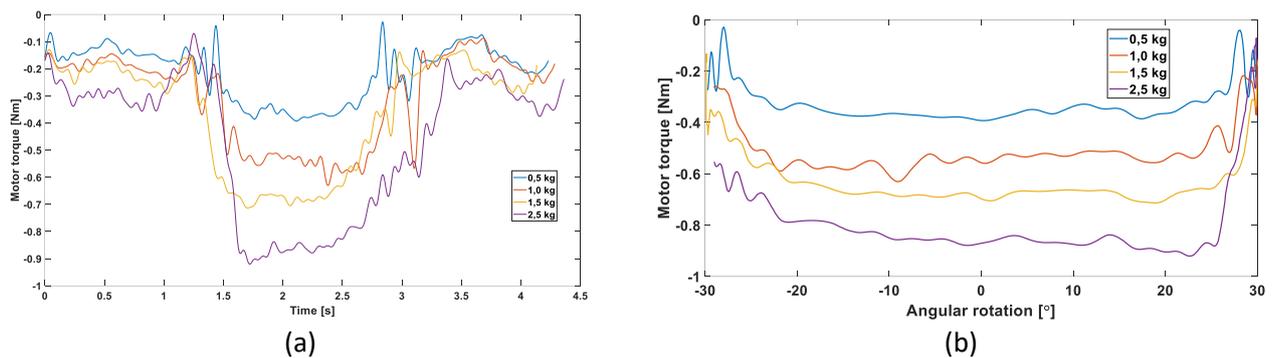

**Figure 16. Test results with the light CADEL prototype for a flexion/bending exercise: (a) right-side motor torque response for different loads vs time, (b) right-side motor torque vs elbow angle rotation.**

Table 3 lists a summary of the characteristics of the various prototypes with the related improvements with indications of the most important specifications that characterize their structural design and operational performance.

As listed in Table 3, main improvements in the evolution of CADEL design have been focused in reducing weight and power consumption to give features for more comfortable usage.

**Table 3 A summary of changes in the CADEL versions**

| Version | CADEL | CADEL.3 | L-CADEL |
|---|---|---|---|
| Year / Location | 2017 / Rome | 2020 / Padova | 2020 / Poitiers |
| Number of cables | 4 | 4 | 2 |
| Number of actuators | 4 | 4 | 3 |
| Number of rings | 2 | 2 | 2 |
| Weight | 2.5 kg | 1 kg | 0.8 kg |
| Power consumption | 3.8 W | 2.66 W | 2.0 W |
| Main improvements | Original design | Inflatable interface<br>Cable elbow guides | Light ring platforms<br>Light-orientable actuators |

## 5 Conclusion

Improvements of CADEL, Cable-Driven ELbow assisting device, are presented with new solutions and mechanical designs for better portability, user comfort and user-oriented operation. A survey of CADEL design is also outlined with the aim to discuss the encountered problems and approaches for the presented improvements. Problems and requirements have been discussed as referring to the improvements that have been achieved in new designs and prototypes in collaborations with different perspectives in particular, the original design is characterised by the conceptual solution that has been augmented in terms of light design, comfort, portability and user-oriented operation with the solution of L-CADEL, mainly developed at Poitiers university, and CADEL.3 mainly developed at Padova University. Main improvements and new solutions can be summarized in light ring platforms equipped with inflatable interface, cable actuation with light-orientable actuators and cable elbow guides to ensure efficient cable tension for elbow motion in bending / extension with arm in sagittal plane. The above aspects have been validated and tested with numerical performance evaluations in lab experiences that will be extended in clinical trial in future work.